\documentclass[runningheads]{llncs}
\usepackage{graphicx}
\usepackage{comment}
\usepackage{amsmath,amssymb} 
\usepackage{color}

\usepackage{mathtools}
\usepackage{bm}
\usepackage{multirow}
\usepackage[table]{xcolor}
\usepackage{enumitem}
\usepackage{pifont}
\newcommand{\cmark}{\ding{51}}
\newcommand{\xmark}{\ding{55}}
\usepackage[pagebackref=true,breaklinks=true,letterpaper=true,colorlinks,bookmarks=false]{hyperref}

\DeclarePairedDelimiter\floor{\lfloor}{\rfloor}

\begin{document}
\pagestyle{headings}
\mainmatter
\def\ECCVSubNumber{4763}  

\title{Fully Convolutional Networks for Continuous Sign Language Recognition} 

%
\author{Ka Leong Cheng\inst{1} \and
Zhaoyang Yang\inst{2} \and
Qifeng Chen\inst{1} \and
Yu-Wing Tai\inst{1,3}
}
%
\authorrunning{K. L. Cheng et al.}
%
\institute{The Hong Kong University of Science and Technology \\
\email{\{klchengad,cqf\}@ust.hk} \and
Tencent \\
\email{yangzhaoyang6@126.com} \and
Kwai Inc. \\
\email{yuwing@gmail.com}}

\maketitle

\begin{abstract}
Continuous sign language recognition (SLR) is a challenging task that requires learning on both spatial and temporal dimensions of signing frame sequences. Most recent work accomplishes this by using CNN and RNN hybrid networks. However, training these networks is generally non-trivial, and most of them fail in learning unseen sequence patterns, causing an unsatisfactory performance for online recognition. In this paper, we propose a fully convolutional network (FCN) for online SLR to concurrently learn spatial and temporal features from weakly annotated video sequences with only sentence-level annotations given. A gloss feature enhancement (GFE) module is introduced in the proposed network to enforce better sequence alignment learning. The proposed network is end-to-end trainable without any pre-training. We conduct experiments on two large scale SLR datasets. Experiments show that our method for continuous SLR is effective and performs well in online recognition.
\end{abstract}

\section{Introduction}
\label{section1}

Sign language is a common communication method for people with disabled hearing. It composes of a variety range of gestures, actions, and even facial emotions. In linguistic terms, a gloss is regarded as the unit of the sign language \cite{article}. To sign a gloss, one may have to complete one or a series of gestures and actions. However, many glosses have very similar gestures and movements because of the richness of the vocabulary in a sign language. Also, because different people have different action speeds, a same signing gloss may have different lengths. Not to mention that different from spoken languages, sign language like ASL \cite{ASL} usually does not have a standard structured grammar. These facts place additional difficulties in solving continuous SLR because it requires the model to be highly capable of learning spatial and temporal information in the signing sequences.

Early work on continuous SLR \cite{10.1007/978-3-319-16178-5_42,article2,6544211} utilizes hand-crafted features followed by Hidden Markov Models (HMMs) \cite{article4,7552950} or Dynamic Time Warping (DTW) \cite{article3} as common practices. More recent approaches achieve state-of-the-art results using CNN and RNN hybrid models \cite{inproceedings2,huang2018videobased,yang2019sfnet}. However, we observe that these hybrid models tend to focus on the sequential order of seen signing sequences in the training data but not the glosses, due to the existence of RNN. So, it is sometimes hard for these trained networks to recognize unseen signing sequences with different sequential patterns. Also, training of these models is generally non-trivial, as most of them require pre-training and incorporate iterative training strategy \cite{inproceedings2}, which greatly lengthens the training process. Furthermore, the robustness of previous models is limited to sentence recognition only; most of the methods fail when the test cases are signing videos of a phrase (sentence fragment) or a paragraph (several sentences). Online recognition requires good recognition responses for partial sentences, but these models usually cannot give correction recognition until the signer finishes all the signing glosses in a sentence. Such limitation in robustness makes online recognition almost impossible for CNN and RNN hybrid models.

In this paper, we propose a fully convolutional network \cite{7298965} for continuous SLR to address these challenges. The proposed network can be trained end-to-end without any pre-training. On top of this, we introduce a GFE module to enhance the representativeness of features. The FCN design enables the proposed network to recognize new unseen signing sentences, or even unseen phrases and paragraphs. We conduct different sets of experiments on two public continuous SLR datasets. The major contribution of this work can be summarized:

\begin{enumerate}
\item We are the first to propose a fully convolutional end-to-end trainable network for continuous SLR. The proposed FCN method models the semantic structure of sign language as glosses instead of sentences. Results show that the proposed network achieves state-of-the-art accuracy on both datasets, compared with other RGB-based methods.
\item The proposed GFE module enforces additional rectified supervision and is jointly trained along with the main stream network. Compared with iterative training, joint training with the GFE module fastens the training process because joint training does not require additional fine-tuning stages.
\item The FCN architecture achieves better adaptability in more complex real-world recognition scenarios, where previous LSTM based methods would almost fail. This attribute makes the proposed network able to do online recognition and is very suitable for real-world deployment applications.
\end{enumerate}

\section{Related Work}
\label{section2}

There are mainly two scenarios in SLR: isolated SLR and continuous SLR. Isolated SLR mainly focuses on the scenario where glosses have been well segmented temporally. Work in the field generally solves the task with methods such as Hidden Markov Models (HMMs) \cite{article6,7532885,article5,5981681,4959905,1640831}, Dynamic Time Warping (DTW) \cite{article7}, and Conditional Random Field (CRF) \cite{5595973,1699159}. As for continuous SLR, the task becomes more difficult as it aims to recognize glosses in the scenarios where no gloss segmentation is available but only sentence-level annotations as a whole. Learning separated individual glosses becomes more difficult in the weakly supervised setting. Many approaches propose to estimate the temporal boundary of different glosses first and then apply isolated SLR techniques and sequence to sequence methods \cite{1004172,5319302} to construct the sentence.

Concerning temporal boundary estimation, Cooper and Bowden \cite{5206647} develop a method to extract similar video regions for inferring alignments in videos by using data mining and head and hand tracking. Farhadi and Forsyth \cite{1640930} also come up with a method that utilizes HMMs to build a discriminative model for estimating the start and end frames of the glosses in video streams with a voting method. Yin et al. \cite{inproceedings3} make further improvements by introducing a weakly supervised metric learning framework to address the inter-signer variation problem in real applications of SLR.

As for sequence to sequence methods, much work follows the framework used in the topic of speech recognition \cite{miao2015eesen,7178778}, handwriting recognition \cite{Liwicki2007ANA,8269951}, and video captioning \cite{wang2018reconstruction}. Specifically, an encoder module is responsible for extracting features in the input video frame sequences, and a CTC module acts as a cost function to learn the ground truth sequences. This framework also shows good performance on continuous SLR, and more recent work applies CNN and RNN hybrid models to infer gloss alignments implicitly \cite{8237594,huang2018videobased,7780825,inproceedings5}. However, RNNs are sometimes more sensitive to the sequential order than the spatial features. As a result, these models tend to learn much about the sequential signing patterns but little about the glosses (words), causing the failure of the recognition for unseen phrases and paragraphs.

\section{Method}
\label{section3}

\begin{figure*}
\begin{center}
\includegraphics[scale=0.4]{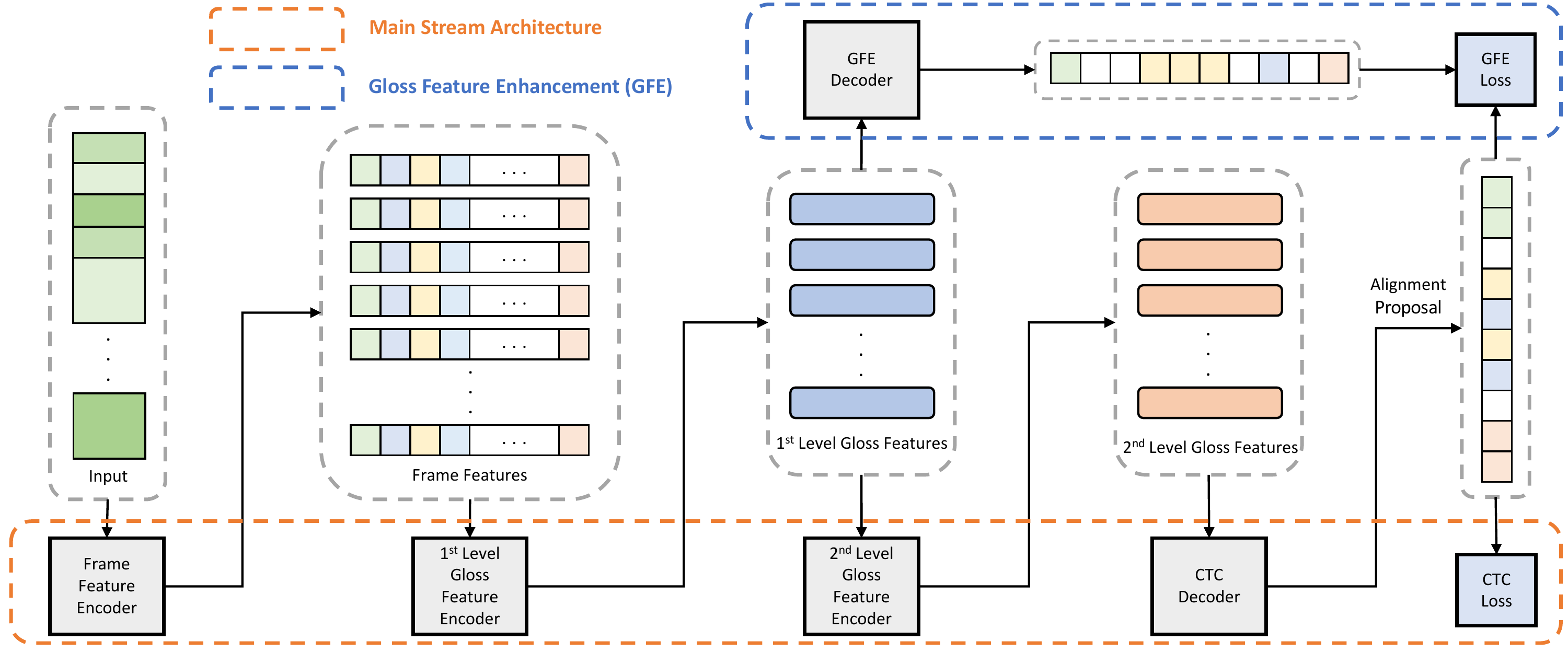}
\end{center}
\caption{Overview of the proposed network. The network is fully convolutional and divides the feature encoding process into two main steps. A GFE module is introduced to enhance the feature learning}
\label{fig:workflow}
\end{figure*}

Formally, the proposed network aims to learn a mapping $H:\mathcal{X} \mapsto \mathcal{Y}$ that can transform an input video frame sequence $\mathcal{X}$ to a target sequence $\mathcal{Y}$. The feature extraction contains two main steps: a frame feature encoder and a two-level gloss feature encoder. On top of them, a gloss feature enhancement (GFE) module is introduced to enhance the feature learning. An overview of the proposed network is shown in Figure \ref{fig:workflow}.

\subsection{Main stream design}
\label{subsection3.1}

\noindent\textbf{Frame feature encoder.} The proposed network first encodes spatial features of the input RGB frames. The frame feature encoder $S$ composes of a convolutional backbone $S_{cnn}$ to extract features in the frames and a global average pooling layer $S_{gap}$ to compress the spatial features into feature vectors. Formally, each signing sequence is a tensor with shape ($t$, $c$, $h$, $w$), where $t$ denotes the length of the sequence, $c$ denotes the number of channels, and $h$, $w$ denotes the height and width of the frames. The process of encoder $S$ can be described as:
\begin{equation}
\{s\}^{t \times f_s} = S(\{x\}^{t \times c \times h \times w}) = S_{gap}(S_{cnn}(\{x\}^{t \times c \times h \times w})).
\end{equation}
The output is of shape $\{s\}^{t \times f_s}$. Note that frame feature encoder treats each frame independently for the frame (spatial) feature learning.

\bigbreak
\noindent\textbf{Two-level gloss feature encoder.} The two-level gloss feature encoder $G$ follows $S$ immediately and aims to encode gloss features. Instead of using LSTM layers, a common practice in temporal feature encoding, we achieve this by using 1D convolutional layers over time dimension. Precisely, the first level encoder $G_1$ consists of 1D-CNNs with a relatively larger filter size. Pooling layers can be used between convolutional layers to increase the window size when needed. Differently, the filter size is relatively smaller for the 1D-CNNs in the second level encoder $G_2$, with no pooling layers used in $G_2$. So, $G_2$ does not change the temporal dimension but only reconsider the contextual information between glosses by taking into account the neighboring glosses.

The overall convolutional process of $G$ can be interpreted as a sliding window on the frame feature vector $\{s\}^{t \times f_s}$ along the time dimension. The sliding window size $l$ and the stride $\delta$ are determined by the accumulated receptive field size and the accumulated stride of 1D-CNNs in $G_1$. Let $\{g\}^{k \times f_{g}}$ and $\{g'\}^{k \times f_{g'}}$ be the output tensor of gloss feature encoder $G_1$ and $G_2$, respectively. The operation of the encoder $G$ can be formulated as:
\begin{equation}
\{g'\}^{k \times f_{g'}} = G(\{s\}^{t \times f_s}) = G_2(G_1(\{s\}^{t \times f_s})) = G_2(\{g\}^{k \times f_{g}}),
\end{equation}
where $k$ is the number of encoded gloss features and can be calculated with:
\begin{equation}
k = \floor{\frac{t - l}{\delta}} + 1.
\end{equation}

The two-level gloss feature encoder takes into account only multiple frames at a time. The window size $l$ should be designed to be around the average length of the signing glosses to ensure good performance during the gloss feature extraction. With a proper window size design, $G_1$ can better model the semantic information of a ``gloss'' in sign language. $G_2$ further considers the gloss neighborhood information to achieve better prediction.

One benefit of our FCN design over previous LSTM design is that it greatly increases the adaptability of recognition, especially for online applications. Our proposed network can provide high-quality recognition on sequences with various length, which is essential in real-world recognition scenarios. We will further discuss the advantages of the FCN design in Section \ref{subsection4.4}.

\bigbreak
\noindent\textbf{CTC decoder.} The Connectionist Temporal Classification (CTC) \cite{inproceedings4} is used as the network decoder. The CTC decoder $D$ aims to decode the encoded gloss feature $\{g'\}^{k \times f_{g'}}$. CTC is an objective function that considers all possible underlying alignments between the input and target sequence. An extra ``blank'' label is added in the prediction space to match the output sequence with the target sequence in temporal dimension. Specifically, we employ a fully connected layer $D_{fc}$ after $G$ to cast the gloss feature dimension from ($k$, $f_{g'}$) to ($k$, $u$) and a Softmax activation to finally transform the gloss feature to the prediction space $\{z\}^{k \times u}$:
\begin{equation}
\{z\}^{k \times u} = D(\{g'\}^{k \times f_{g'}}) = softmax(D_{fc}(\{g'\}^{k \times f_{g'}})),
\end{equation}
where $v$ is the vocabulary size and $u=v+1$ is the size of each output with the extra ``blank'' added.

With normalized probabilities $\{z\}^{k \times u}$, the output alignment $\bm{\pi}$ can then be generated by taking the label with maximum likelihood at every decoding step. The final recognition result $\bm{y}$ is obtained by using the many-to-one function $\mathcal{B}$ introduced in CTC to remove repeated and blank predictions in $\bm{\pi}$. The CTC objective function is defined as the negative log-likelihood of all possible alignments matched to the ground truth:
\begin{equation}
\mathcal{L}_{ctc}(\bm{x},\bm{y}) = -log\ p(\bm{y}|\bm{x}).
\end{equation}

With the additional $l_2$ regularizer $\mathcal{L}_{reg}$ on the network parameters $\bm{W}$, the objective function of the main stream of the network $\mathcal{L}_{main}$ is defined as:
\begin{equation}
\begin{split}
\mathcal{L}_{main} &= \mathcal{L}_{ctc} + \lambda_1\mathcal{L}_{reg} \\
&= \frac{1}{|\mathcal{S}|} \sum\limits_{(\bm{x},\bm{y})\in\mathcal{S}}\mathcal{L}_{ctc}(\bm{x},\bm{y}) + \lambda_1||\bm{W}||^2,
\end{split}
\end{equation}
where $\mathcal{S}$ is the sample space, and $\lambda_1$ is the weight factor of the regularizer.

\subsection{Gloss feature enhancement}
\label{subsection3.2}

\begin{figure*}[t]
\begin{center}
\includegraphics[scale=0.72]{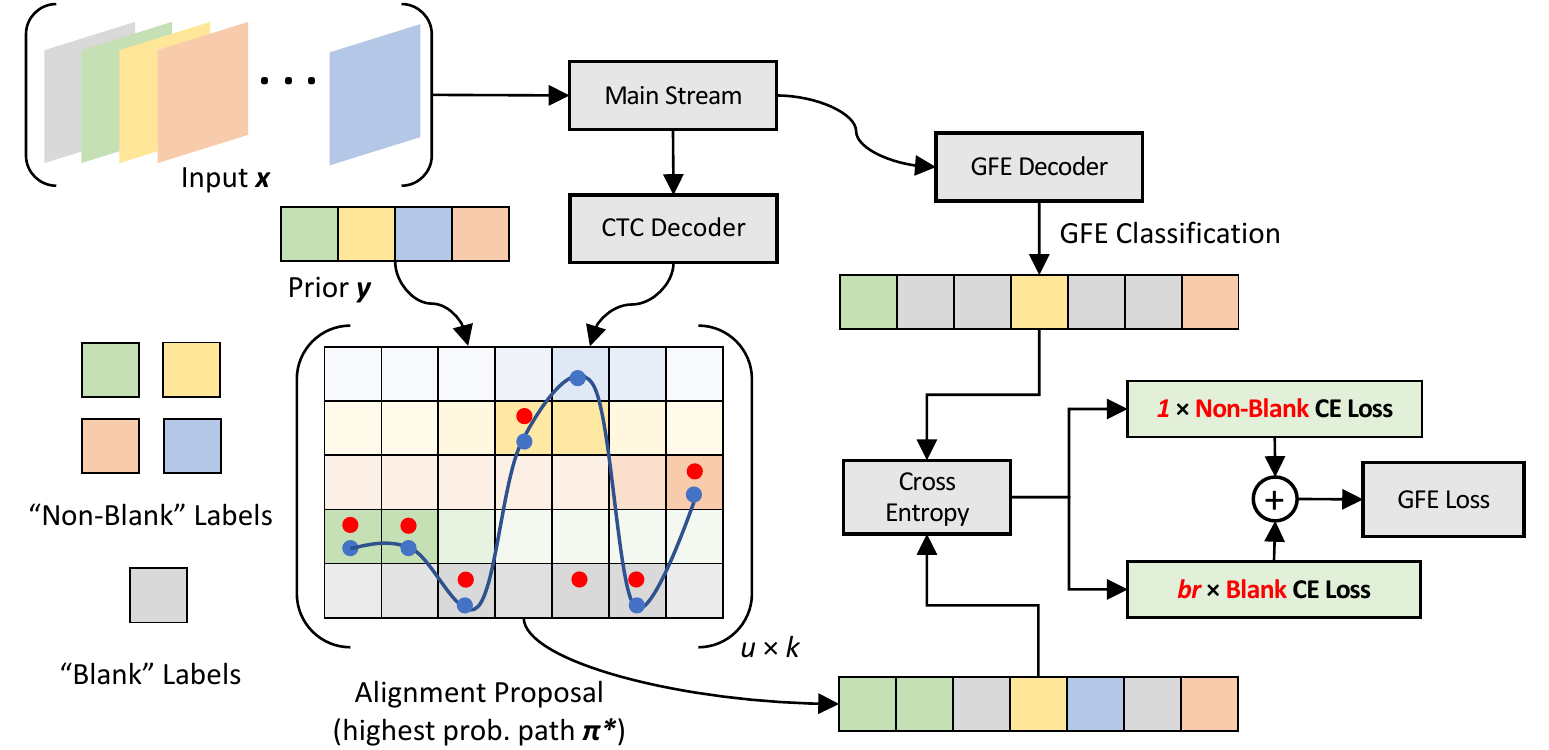}
\end{center}
\caption{The GFE module. The red dots in the prediction map are network outputs, while the blue ones are alignment proposals. The proposal rectifies the false predictions in the output to match the ground truth}
\label{fig:gfe}
\end{figure*}  

The main stream of the network has mainly two tasks: (1) alignment inference and (2) gloss prediction. The performance highly depends on how well the network can generalize on glosses, as they are the unit of sign language. Therefore, it is essential to improve the quality of gloss features. Previous methods generally achieve this by incorporating iterative training strategies \cite{inproceedings2,Koller2017ReSignRE,Pu_2019_CVPR,Cui2019ADN}. They first break training into several stages and then gradually refine feature extraction as each stage processed. However, with this strategy, whenever the training is switched to another stage, the network needs to first gradually adapt to a different objective, which greatly lengthens the number of training epochs and reduces the training efficiency. Moreover, the supervision used in some methods is generally the output of the network, which may contain some false predictions that can further reduce learning efficiency and limit the effectiveness of the refinement. 

To remedy these problems, we propose the GFE module. The GFE module uses rectified supervision and is jointly trained with the main stream of the network, so it can improve the the main stream network performance on the line. We illustrate the idea of the GFE module in Figure \ref{fig:gfe}.

\bigbreak
\noindent\textbf{Alignment proposal.} High-quality supervision can significantly improve the effectiveness of feature enhancement. Similar to \cite{Cui2019ADN}, we make use of the network prediction map to find a better alignment proposal as the supervision. Specifically, given an input sequence, the CTC decoder $D$ generates a prediction map $\{z\}^{k \times u}$, which is the probability of emissions in each decoding step. Let $\pi^*$ denote the element in the alignment proposal. The alignment proposal $\bm{\pi^*} = \{\pi^*\}^k$ used in the GFE module can then be generated by searching the alignment proposal with the highest probability that can be matched to the ground truth sequence:
\begin{equation}
\bm{\pi^*} = \mathop{\arg\max}_{\bm{\pi}\in\mathcal{B}^{-1}(\bm{y})} p(\bm{\pi}|\bm{x}),
\end{equation}
where $\mathcal{B}^{-1}$ is the inverse function of $\mathcal{B}$. Hence, the alignment proposal is guaranteed to be a matched alignment of the ground truth sequence. Each $\pi^*$ in $\bm{\pi^*}$ can be paired with a first level gloss feature vector $g$ at the corresponding time step in $\bm{g}$, which gives a pair of learning sample $(g, \pi^*) \in \mathcal{V}$.

\bigbreak
\noindent\textbf{Joint training with weighted cross-entropy.} To use the learning pairs in $\mathcal{V}$ as enhancement supervision, we add a fully connected layer $F_{fc}$ followed by a Softmax activation after $G_1$. When joint training with the GFE module, gradients along this addition branch only propagate back to $F$ and $G_1$ to enhance the frame and gloss feature learning. Formally, the GFE module contains a GFE decoder $F$, that takes $\bm{g} = \{g\}^{k \times f_g}$, the output vector of $G_1$, as inputs, and outputs the predicted gloss sequence in prediction space $\bm{\hat\pi^*} = \{\hat\pi^*\}^{k \times u}$:
\begin{equation}
\{\hat\pi^*\}^{k \times u} = F(\{g\}^{k \times f_g}) = softmax(F_{fc}(\{g\}^{k \times f_g})).
\end{equation}

It is intuitive to train the gloss feature enhancement branch with cross-entropy loss. However, it is common that most of the label $\pi^*$ in $\bm{\pi^*}$ is ``blank'' as $k$ is generally much bigger than the number of glosses in the ground truth, causing the imbalance of samples in $\mathcal{V}$. The sample imbalance may limit the effectiveness of training for the GFE module. Therefore, we introduce a balance ratio to decrease the loss from ``blank'' labels. The balance ratio is defined as the proportion of ``non-blank'' labels in the given proposal $\bm{\pi^*}$:
\begin{equation}
br = \frac{\#non\textnormal{-}blank}{\#total}.
\end{equation}

For every $(g, \pi^*) \in \mathcal{V}$, we re-scale the cross-entropy loss to obtain the GFE loss, where the scaling factor $w_i$ equals to $br$ if it is blank label ($i=u$), otherwise $w_i$ equals to $1$:
\begin{equation}
\begin{split}
\mathcal{L}_{gfe}(g, \pi^*) = -\frac{1}{u}\sum\limits_{i=1}^{u} w_ilog\ p(\pi=\pi^*_i|g).
\end{split}
\end{equation}
With the GFE module, the overall objective of the proposed network $\mathcal{L}$ becomes:
\begin{equation}
\begin{split}
\mathcal{L} =\ &\mathcal{L}_{main} + \lambda_2\mathcal{L}_{gfe} \\
=\ &\frac{1}{|\mathcal{S}|} \sum\limits_{(\bm{x},\bm{y})\in\mathcal{S}}\mathcal{L}_{ctc}(\bm{x},\bm{y})\ +\\
&\frac{\lambda_2}{|\mathcal{V}|} \sum\limits_{(g, \pi^*)\in\mathcal{V}}\mathcal{L}_{gfe}(g, \pi^*) + \lambda_1||\bm{W}||^2,
\end{split}
\end{equation}
where $\lambda_2$ is the weight factor for the GFE module. Note that the network objective is unified with joint training, so the training process is more efficient.

\section{Experiments}
\label{section4}

We conduct experiments on the Chinese Sign Language (CSL) dataset \cite{huang2018videobased} and the RWTH-PHOENIX-Weather-2014 (RWTH) dataset \cite{article2}. We detail the experimental setup, results, ablation studies, and online recognition in this section.

\subsection{Experimental setup}
\label{subsection4.1}

\noindent\textbf{Dataset.} The RWTH-PHOENIX-Weather-2014 (RWTH) dataset is recorded from a public weather broadcast television station in Germany. All signers wear dark clothes and perform sign languages in front of a clean background. There are $6,841$ different sentences signed by $9$ different signers (around $80,000$ glosses with a vocabulary of size $1,232$). All videos are pre-processed to a resolution of $210 \times 260$ and $25$ frames per second (FPS). The dataset is officially split with $5,672$ training samples, $540$ validation samples, and $629$ testing samples.

The Chinese Sign Language (CSL) dataset contains $100$ sentences, each being signed for $5$ times by $50$ signers (in total $25,000$ videos). Videos are shoot using a Microsoft Kinect camera with a resolution of $1280 \times 720$ and a frame rate of $30$ FPS. The vocabulary size is relatively small ($178$); however, the dataset is richer in performance diversity, since signers wear different clothes and sign with different speeds and action ranges. With no official split given, we divide the dataset into a training set of $20,000$ samples and a testing set of $5,000$ samples and ensure that the sentences in the training and testing sets are the same, but the signers are different.

\bigbreak
\noindent\textbf{Evaluation metric.} We use word error rate (WER), which is the metric commonly used in continuous SLR, to evaluate the performance of recognition:
\begin{equation}
WER(H(\bm{x}), \bm{y})) = \frac{\#ins + \#del + \#sub}{\#labels\ in\ \bm{y}}.
\end{equation}
We treat a Chinese character as a word during evaluation for the CSL dataset.

\begin{figure*}[t]
\begin{center}
\includegraphics[scale=0.53]{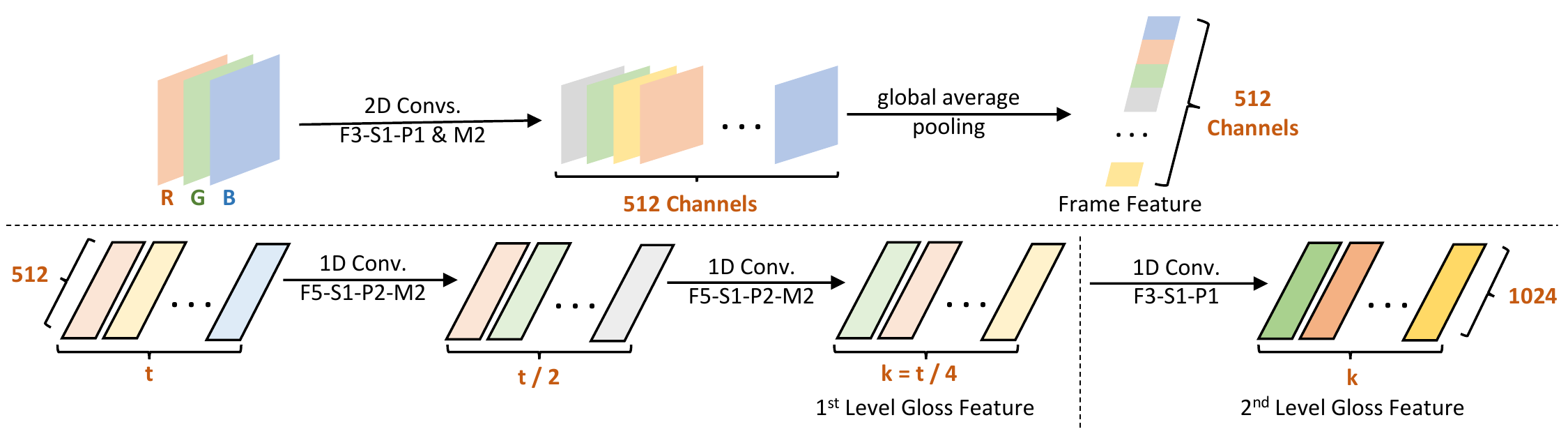}
\end{center}
\caption{The setting of the proposed main stream network. F, M refer to the filter size of convolution and max-pooling, respectively. S, P refer to the stride and padding size of convolution, respectively. Numbers aside are their actual size}
\label{fig:detail}
\end{figure*}

\bigbreak
\noindent\textbf{Implementation details.} The main stream network setting used in our experiments is shown in Figure \ref{fig:detail}. For $S_{cnn}$, the channel number gradually increases in this pattern: 3-32-64-64-128-128-256-256-512-512. F3-S1-P1 is used in each layer, and an additional M2 is added if the channel number increases. $S_{gap}$ does global average pooling on each channel, so each frame is encoded as an array with a length of $512$. For encoder $G_1$, two F5-S1-P2-M2 layers are used, and the channel number remains unchanged as 512. For encoder $G_2$, one F3-S1-P1 layer is used, and the channel number increases to 1024. Both fully connected layers $D_{fc}$ and $F_{fc}$ in the main stream and the GFE module cast the input channel number to $u$, the number of vocabulary size plus one blank label.

Batch normalization \cite{ioffe2015batch} is added after every convolutional layer to accelerate training. The input resolution of the network is $224 \times 224$. The window size in the first level gloss feature encoder is set to be $16$ (about $0.5$-$0.6$ seconds), which is the average time needed for completing a gloss, and the stride of the window is set to be $4$. The second level gloss feature encoder further considers $3$ adjacent gloss features for better prediction.

We use Adam \cite{kingma2014adam} optimizer for training. We set the initial learning rate to be $10^{-4}$. The weight factor $\lambda_1$ and $\lambda_2$ are empirically set to be $10^{-4}$ and $0.05$, respectively. For the RWTH dataset, we train the proposed network for $80$ epochs and halve learning rate at epoch $40$ and $60$. For the CSL dataset, the network is trained for $60$ epochs, with the learning rate reduced by half at epoch $30$ and $45$. For data augmentation, all frames are first resized to $256 \times 256$ and then randomly cropped to fit the input shape. We also do temporal augmentation by first scaling up the sequence by $+20\%$ and then by $-20\%$. Joint training with the GFE module is activated after epoch $15$ for RWTH and epoch $10$ for CSL, which are chosen through experiments to avoid unreliable alignment proposal at the initial optimization stage. The alignment proposals in the GFE module are updated every $10$ epochs. When updating the proposal, temporal augmentation is disabled.

\subsection{Results}
\label{subsection4.2}

We give a thorough comparison between the proposed network and previous RGB-based methods on both datasets. The results of previous methods are collected from their original papers. Please note that we mainly focus on online recognition in SLR, where the inputs are usually RGB video frames. Hence, we only compare our results with previous methods that use solely RGB modality.

\begin{table}[t]
    \begin{minipage}{.47\linewidth}
      \caption{Result comparison on CSL}
      \begin{center}
      \begin{tabular}{p{90pt}|p{35pt}}
      \hline
      \hfil\textbf{Methods} & \hfil\textbf{WER}\\
      \hline
      \hfil\textbf{DTW-HMM}~\cite{article3} & \hfil 28.4\\
      \hfil\textbf{LSTM}~\cite{venugopalan2014translating} & \hfil 26.4\\
      \hfil\textbf{S2VT}~\cite{venugopalan2015sequence} & \hfil 25.5\\
      \hfil\textbf{LSTM-A}~\cite{yao2015describing} & \hfil 24.3\\
      \hfil\textbf{LSTM-E}~\cite{pan2015jointly} & \hfil 23.2\\
      \hfil\textbf{HAN}~\cite{article4} & \hfil 20.7\\
      \hfil\textbf{LS-HAN}~\cite{huang2018videobased} & \hfil 17.3\\
      \hfil\textbf{SubUNet}~\cite{8237594} & \hfil 11.0\\
      \hfil\textbf{HLSTM}~\cite{Guo2018HierarchicalLF} & \hfil 7.6\\
      \hfil\textbf{HLSTM-attn}~\cite{Guo2018HierarchicalLF} & \hfil 7.1\\
      \hfil\textbf{Align-iOpt}~\cite{Pu_2019_CVPR} & \hfil 6.1\\
      \hfil\textbf{SF-Net}~\cite{yang2019sfnet} & \hfil 3.8\\
      \hline
      \hfil\textbf{Ours} & \hfil\textbf{3.0}\\
      \hline
      \end{tabular}
      \end{center}
      \label{tab:csl}
    \end{minipage}%
    \begin{minipage}{.01\linewidth}
    \
    \end{minipage}
    \begin{minipage}{.51\linewidth}
      \caption{Result comparison on RWTH}
      \begin{center}
      \begin{tabular}{p{95pt}|p{30pt}p{30pt}}
      \hline
      \multirow{2}{95pt}{\centering\textbf{Methods}} & \multicolumn{2}{c}{\textbf{WER}}\\\cline{2-3}
      & \hfil\textbf{Dev} & \hfil\textbf{Test}\\
      \hline
      \hfil\textbf{Koller et al.}~\cite{article2} & \hfil 57.3 & \hfil 55.6\\
      \hfil\textbf{Deep Hand}~\cite{7780781} & \hfil 47.1 & \hfil 45.1\\
      \hfil\textbf{Deep Sign}~\cite{inproceedings7} & \hfil 38.3 & \hfil 38.8\\
      \hfil\textbf{SubUNet}~\cite{8237594} & \hfil 40.8 & \hfil 40.7\\
      \hfil\textbf{Cui et al.}~\cite{inproceedings2} & \hfil 39.4 & \hfil 38.7\\
      \hfil\textbf{LS-HAN}~\cite{huang2018videobased} & \hfil - & \hfil 38.3\\
      \hfil\textbf{Align-iOpt}~\cite{Pu_2019_CVPR}& \hfil 37.1 & \hfil 36.7\\
      \hfil\textbf{SF-Net}~\cite{yang2019sfnet} & \hfil 38.0 & \hfil 38.1\\
      \hfil\textbf{Re-Sign}~\cite{Koller2017ReSignRE}& \hfil 27.1 & \hfil 26.8\\
      \hfil\textbf{STMC (RGB)}~\cite{Zhou2020SpatialTemporalMN} & \hfil 25.0 & \hfil -\\
      \hfil\textbf{Cui et al. (RGB)}~\cite{Cui2019ADN} & \hfil 23.8 & \hfil 24.4\\
      \hline
      \hfil\textbf{Ours} & \hfil\textbf{23.7} & \hfil\textbf{23.9}\\
      \hline
      \end{tabular}
      \end{center}
      \label{tab:rwth}
    \end{minipage} 
\end{table}

\begin{figure*}[t]
\begin{center}
\includegraphics[scale=0.34]{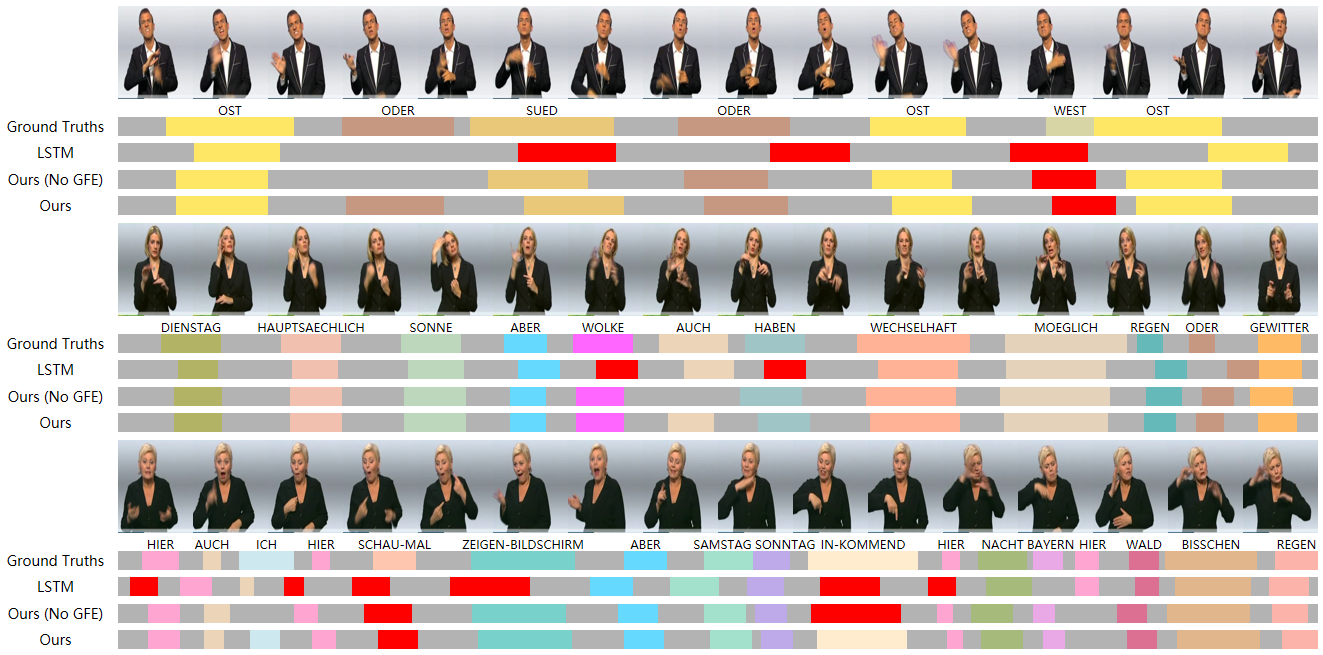}
\end{center}
\caption{Sample outputs for different network settings. Wrong recognitions (except deletion errors) are highlighted in red. Ground truths are manually aligned}
\label{fig:alignment}
\end{figure*}

Results on the CSL dataset and the RWTH dataset are shown in Table~\ref{tab:csl} and Table~\ref{tab:rwth}, respectively. We see that the proposed network achieves state-of-the-art performance on both datasets for RGB-based methods. The best result achieves $3.0\%$ for the CSL dataset. For the RWTH dataset, our model reports $23.7\%$ for the development set and $23.9\%$ for the testing set. We also test the performance on the recognition partition (without translation) of the RWTH-PHOENIX Weather 2014 \textbf{T} dataset \cite{inproceedings}. Our WER on the development set and testing set are $23.3\%$ and $25.1\%$, respectively. These results illustrate the effectiveness of the proposed network.

To further demonstrate the effectiveness of our methods, we show some sample outputs in Figure \ref{fig:alignment} to compare the recognition quality of different network settings. We observe that in the LSTM setting, models recognize the identical glosses (such as ``OST'' in sample $1$ and ``HIER'' in sample $3$) in a sentence as different words, and that errors usually occur adjacently in the sentences. In contrast, the proposed network produces consistent results for the identical glosses in a sentence, and errors are usually isolated. The observations infer that the LSTM based methods tend to learn robust sequential information, while the proposed network focuses on learning strong gloss features. So, we claim that the proposed network has a better generalization capability, because identical glosses are consistently classified, and errors do not have significant effects on neighboring glosses. On top of that, the GFE module further improves the performance by rectifying wrong recognition (such as ``IN-KOMMEND'' in sample $3$), finding missing recognition (such as ``AUCH'' in sample $2$), and adjusting alignments. More qualitative comparison can be found in the supplementary material.

\subsection{Ablation studies}
\label{subsection4.3}

In this section, we present further ablation studies to demonstrate the effectiveness of our method.

\bigbreak
\noindent\textbf{Temporal feature encoder.} We first conduct a set of experiments to compare network performance with different temporal feature encoder designs. We test $6$ different design combinations for the temporal feature encoder. For notation, \textbf{None} means no architecture; \textbf{LSTM} or \textbf{BiLSTM} means 1 LSTM or BiLSTM layer with 512 hidden states, respectively; \textbf{1D-CNN} means two F5-S1-P2-M2 1DConvs for the $1^{st}$ level or one F3-S1-P1 1DConv for the $2^{nd}$ level. We show results on the testing set of the RWTH and CSL datasets in Table \ref{tab:CNN-LSTM}. Note that in this set of experiments, the GFE module is not activated.

\begin{table}[b]
    \begin{minipage}{.55\linewidth}
    \caption{Network performance with different temporal feature encoder design}
    \begin{center}
    \begin{tabular}{c|c|p{40pt} p{30pt}}
    \hline
    \multirow{2}{50pt}{\centering\textbf{$1^{st}$ level}} & \multirow{2}{50pt}{\centering\textbf{$2^{nd}$ level}} & \multicolumn{2}{c}{\hfil\textbf{WER}} \\\cline{3-4}
    & & \hfil\textbf{RWTH} & \hfil\textbf{CSL} \\\hline
    \hfil\textbf{None} & \hfil\textbf{1D-CNN} & \hfil 60.5 & \hfil 23.3 \\
    \hfil\textbf{1D-CNN} & \hfil\textbf{None} & \hfil 42.1 & \hfil 10.4 \\
    \hfil\textbf{LSTM} & \hfil\textbf{1D-CNN} & \hfil 32.1 & \hfil 10.8 \\
    \hfil\textbf{LSTM} & \hfil\textbf{BiLSTM} &  \hfil 30.8 & \hfil 3.6 \\
    \hfil\textbf{1D-CNN} & \hfil\textbf{BiLSTM} & \hfil 26.5 & \hfil\textbf{3.4} \\
    \hfil\textbf{1D-CNN} & \hfil\textbf{1D-CNN} & \hfil\textbf{26.0} & \hfil 8.2 \\
    \hline
    \end{tabular}
    \end{center}
    \label{tab:CNN-LSTM}
    \end{minipage}
    \begin{minipage}{.02\linewidth}
    \end{minipage} 
    \begin{minipage}{.42\linewidth}
    \caption{Network performance in different GFE module settings, br refers to the balance ratio}
    \begin{center}
    \begin{tabular}{p{45pt}|p{25pt} p{25pt} p{30pt}}
    \hline
    \hfil\textbf{\ } & \hfil\textbf{GFE} & \hfil\textbf{br} & \hfil\textbf{WER}\\
    \hline
    \multirow{3}{45pt}{\centering\textbf{RWTH}} 
    & \hfil\xmark & \hfil\xmark & \hfil 26.0\\
    & \hfil\cmark & \hfil\xmark & \hfil 25.4\\
    & \hfil\cmark & \hfil\cmark & \hfil\textbf{23.9}\\
    \hline
    \multirow{3}{45pt}{\centering\textbf{CSL}}  
    & \hfil\xmark & \hfil\xmark & \hfil 8.2\\
    & \hfil\cmark & \hfil\xmark & \hfil 4.5\\
    & \hfil\cmark & \hfil\cmark & \hfil\textbf{3.0}\\
    \hline
    \end{tabular}
    \end{center}
    \label{tab:GFE}
    \end{minipage}
\end{table}

We see that both levels of gloss feature encoder are essential for recognition as WER raises significantly when either of them is absent. CNN-based designs consistently outperform their LSTM counterparts for the RWTH dataset, but networks with BiLSTM give the best results for the CSL dataset. We should be reminded of the differences between the two datasets. The CSL dataset has richer diversity in the spatial dimension (such as different cloth colors) than in the temporal term. In other words, it is easier for the network to learn the temporal features than the spatial features in the CSL dataset. Also, different from that all the testing sentences are unseen in the RWTH training set, all the testing sentences in the CSL dataset are already seen in the training set, but just signed by different people. The working nature of 1D-CNN and LSTM is also different. The LSTM based method has direct access to sentence-level information, since LSTM layers have access to all the frame information. While the FCN method has less sentence-level supervision (only indirect access through the CTC decoding function), as the FCN model uses only a fixed number of frames to predict a gloss at each time step with 1D-CNNs.

Therefore, we claim that LSTM based methods tend to ``remember'' all the signing sequences in the training set instead of trying to learn the glosses independently. When testing the sentences in the CSL dataset, whose sequential sentences are the same as the training samples, the strong sequential information learned by LSTM based methods is significant and helpful, causing a relatively low reported WER. While for our FCN method, it is hard to fully extract the spatial and temporal features with weak supervision without substantial sentence-level information. Thus, it is essential to have a GFE module for feature enhancement, and the full version of our proposed network outperforms all the LSTM based methods for both datasets.

\bigbreak
\noindent\textbf{GFE module.} We conduct a set of experiments to investigate the effectiveness of the GFE module. We test different settings on both the RWTH and CSL datasets, including learning without the GFE module, with the GFE module but without balance ratio, and with the GFE module and balance ratio. Results on the testing sets are shown in Table \ref{tab:GFE}.

We see that when the balance ratio is used, the GFE module can significantly fine-tune the features and improve the performances accordingly for both datasets. The GFE module with balance ratio improves the testing WER for the RWTH dataset by $2.1\%$; the improvement is more prominent for the CSL dataset by $5.2\%$. The difference of improvement is because the CSL dataset has much richer diversity in the spatial dimension than the RWTH dataset, making the spatial features in the CSL dataset more difficult to be learned without the GFE module.

\begin{figure*}[t]
\begin{center}
\includegraphics[scale=0.52]{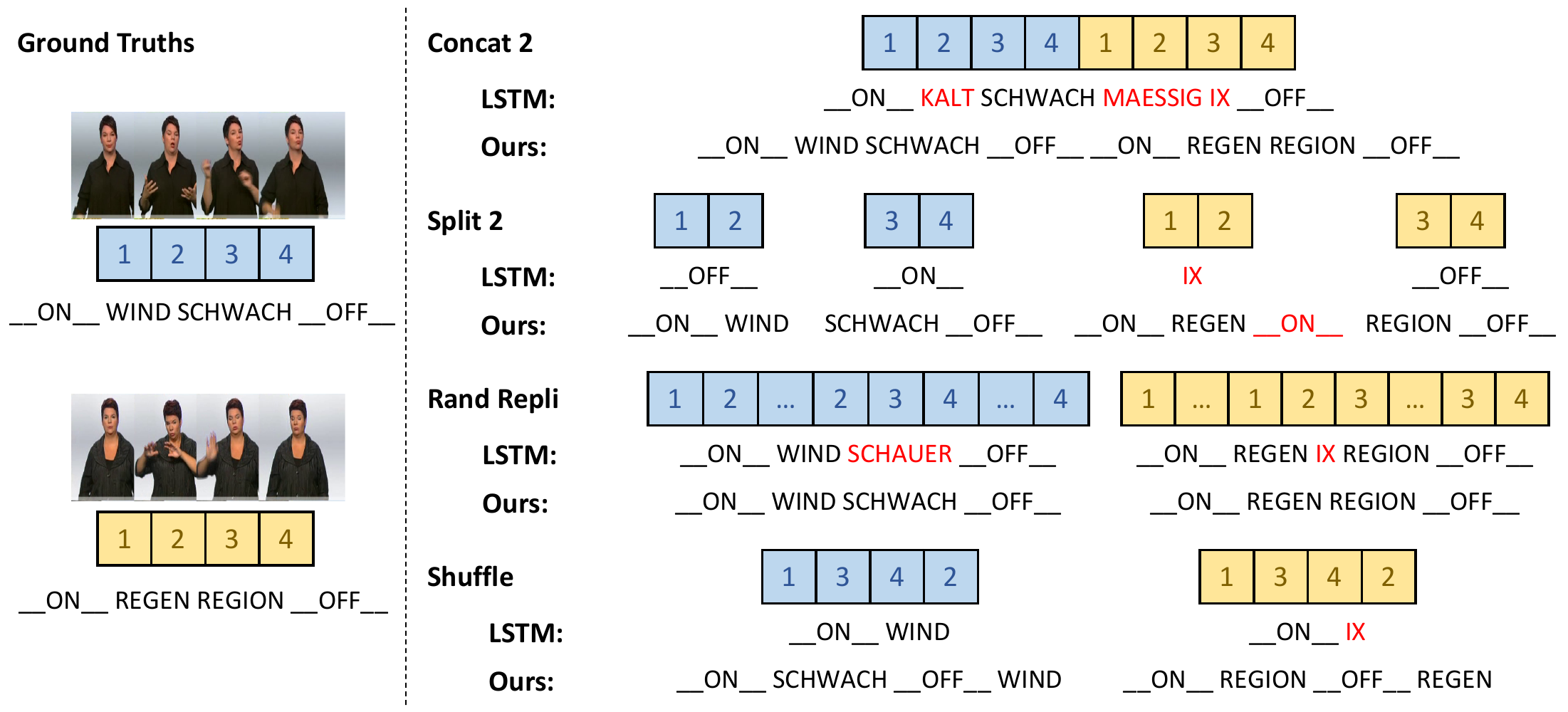}
\end{center}
\caption{Two samples are chosen for illustrating four different simulating scenarios for real-world recognition. Given that both the LSTM based method and the proposed network can recognize the original samples correctly, in all the simulating scenarios, the LSTM based method provides many false predictions while ours can preserve its accuracy. Errors are marked in red}
\label{fig:online-explain}
\end{figure*}

\begin{table}[b]
    \caption{Network performances in different real-world simulating scenarios on RWTH}
    \begin{center}
    \begin{tabular}{p{60pt}|p{35pt} p{35pt} p{35pt} p{35pt}}
    \hline
    \multirow{2}{60pt}{\centering\textbf{Setups}} & \multicolumn{2}{c}{\hfil\textbf{Dev}} & \multicolumn{2}{c}{\hfil\textbf{Test}}\\\cline{2-5}
    & \hfil{ LSTM } & \hfil{ Ours } & \hfil{ LSTM } & \hfil{ Ours }\\
    \hline
    \hfil\textbf{Original} & \hfil 31.7 & \hfil 23.7 & \hfil 30.8 & \hfil 23.9\\
    \hfil\textbf{Split 2} & \hfil 45.6 & \hfil 26.6 & \hfil 42.3 & \hfil 26.4\\
    \hfil\textbf{Split 3} & \hfil 50.2 & \hfil 28.0 & \hfil 45.3 & \hfil 27.6\\
    \hfil\textbf{Concat 2} & \hfil 40.5 & \hfil 24.3 & \hfil 41.1 & \hfil 24.9\\
    \hfil\textbf{Concat 3} & \hfil 46.0 & \hfil 24.8 & \hfil 45.7 & \hfil 25.0\\
    \hfil\textbf{Concat All} & \hfil - & \hfil 25.5 & \hfil - & \hfil 25.3\\
    \hfil\textbf{Rand Repli} & \hfil 39.1 & \hfil 24.6 & \hfil 39.5 & \hfil 24.6\\
    \hfil\textbf{Shuffle} & \hfil 58.9 & \hfil 27.3 & \hfil 55.2 & \hfil 28.1\\
    \hline
    \end{tabular}
    \end{center}
    \label{tab:Online}
\end{table}

\subsection{Online recognition} 
\label{subsection4.4}

We mention in Section \ref{subsection4.3} that the proposed network has weaker supervision on sentence information. The FCN method focuses more on glosses rather than sentences, which directs us for some interesting experiments with different setups.

\bigbreak
\noindent\textbf{Simulating experiments.} For recognition in the real world, it is natural to consider the following three scenarios: (1) Signers sign several sentences at a time. (2) Signers sign only a phrase at a time. (3) Signers may pause for some while (stutters of actions) in the middle of signing. Accordingly, as shown in Figure \ref{fig:online-explain}, we design three types of experiments which are conducted on the RWTH dataset to simulate online recognition: (1) concatenate multiple samples for a new sample (numbers after \textbf{Concat} indicate the number of samples being concatenated); (2) evenly split a sample into multiple samples (numbers after \textbf{Split} indicate the number of equal segments); (3) randomly select $5$ frames in each sample and replace them with $12$ replications in place (\textbf{Rand Repli} means random replication). 

The LSTM based method may have too strong supervision of the sequential order, making it sensitive to gloss order. But one advantage of the proposed network is that the FCN model learns the glosses independently, so it is more robust to order-independent representations. To show this advantage, we may experimentally by shuffling the glosses in the testing samples. However, given no isolated annotations for individual glosses, we cannot manually construct ``new'' sentences with different signing orders by random shuffling. To mimic the shuffling idea, our fourth experiment is an imperfect but reasonable gloss shuffling experiment, as shown as \textbf{Shuffle} in Figure \ref{fig:online-explain}. We first temporally segment the input into two equal parts and insert one into another. 

The results of the four experiments are shown in Table \ref{tab:Online}. It is observed that the performance of the LSTM based network degrades dramatically in all these four types of simulating scenarios. On the contrary, the proposed network shows consistent overall performance across different scenarios. We only observe additional minor errors in the output steps where samples are combined or split due to the action inconsistencies introduced in boundary places.

\bigbreak
\noindent\textbf{Discussion.} Considering the nature of the FCN design, the results further inspire us that even the proposed network is continuously being fed only a few frames that are needed (adequate) to infer the output, it can still combine all the intermediate outputs to give the same final recognition result. We use this technique to test for the \textbf{Concat All} scenario, where all testing samples are concatenated together as one large sample. Unfortunately, the LSTM based model fails to provide any result for the \textbf{Concat All} scenario, as the memory capacity limits the network to take such a large sample as an input.

All the results Table \ref{tab:Online} indicate that the proposed network has more generalization capability and better flexibility for recognition in complex real-world recognition scenarios. The FCN design enables the proposed network to significantly reduce the memory usage for recognition. Meanwhile, indicated by the results in the \textbf{Split} and \textbf{Concat} scenarios, besides recognizing signing sentences, our method gives accurate recognition results for signing phrases and paragraphs. We can further conclude from the results in the \textbf{Split} scenarios that there is no need to wait for the arrival of all the signing glosses during the recognition process, as accurate intermediate (partial) recognition results can be given whenever adequate frames are available to the proposed network. With this great property, our method can provide intermediate results word by word along time, which is very friendly from a human-computer interaction perspective for SLR users. These properties make the proposed network have a promising application prospect for online recognition. A visual demonstration is shown in the supplementary demo video.

\section{Conclusions}
\label{section5}

In this paper, we are the first to propose a fully convolutional network that can be trained end-to-end without pre-training for continuous SLR. A jointly trained GFE module is introduced to enhance the representativeness of features. Experimental results show that the proposed network achieves state-of-the-art performance on benchmark datasets with RGB-based methods. For recognition in real-world scenarios where the LSTM based network mostly fails, the proposed network reaches consistent performance and shows many great advantageous properties. These advantages make our proposed network robust enough to do online recognition.

One possible future research direction for continuous SLR is to strengthen the supervision by using the fact that some glosses are combinations of letter signs; however, this may require additional labeling pre-processing and professional knowledge in sign language. Also, the better gloss recognition accuracy obtained by the proposed network may have a good research prospect in sign language translation (SLT). Furthermore, we hope the proposed network can inspire future studies on sequence recognition tasks to investigate FCN architecture as an alternative to LSTM based models, especially for those tasks with limited data for training.

\clearpage
%
%
\bibliographystyle{splncs04}
\bibliography{egbib}
\end{document}